\documentclass[letterpaper, 10 pt, conference]{ieeeconf}  

\IEEEoverridecommandlockouts                              
\usepackage[linesnumbered,ruled,vlined]{algorithm2e}                      
\usepackage{amsmath,amssymb,amsfonts}
\usepackage{algorithmic}
\usepackage{lmodern}
\usepackage{mathtools}
\usepackage{url}

\usepackage{breakurl}
\usepackage[breaklinks]{hyperref}
\usepackage{multirow}
\usepackage{booktabs}
\usepackage{xcolor}
\usepackage{tabularray}
\usepackage{rotating}
\usepackage{tabularray}
\usepackage{makecell}
\DeclareMathOperator*{\argmin}{arg\,min}

\title{\LARGE \bf
Multimodal Manoeuvre and Trajectory Prediction for Automated Driving on Highways Using Transformer Networks
}

\author{Sajjad Mozaffari, Mreza Alipour Sormoli, Konstantinos Koufos, and Mehrdad Dianati
\thanks{This work was part of the Hi-Drive project. The project has received funding from the European Union's Horizon 2020 research and innovation programme under grant agreement No. 101006664.}
\thanks{S. Mozaffari, M. Alipour Sormoli, K. Koufos are with WMG, University of Warwick, Coventry, U.K. (e-mails: sajjad.mozaffari, mreza.alipour, konstantinos.koufos@warwick.ac.uk)
}
\thanks{M. Dianati is with School of Electronics, Electrical Engineering and Computer Science Queen’s University of Belfast. (e-mail: m.dianati@qub.ac.uk)
}
}
\begin{document}

\maketitle
\thispagestyle{plain}
\pagestyle{plain}


\begin{abstract}

Predicting the behaviour (i.e., manoeuvre/trajectory) of other road users, including vehicles, is critical for the safe and efficient operation of autonomous vehicles (AVs), a.k.a., automated driving systems (ADSs). Due to the uncertain future behaviour of vehicles, multiple future behaviour modes are often plausible for a vehicle in a given driving scene. Therefore, multimodal prediction can provide richer information than single-mode prediction, enabling AVs to perform a better risk assessment. To this end, we propose a novel multimodal prediction framework that can predict multiple plausible behaviour modes and their likelihoods. The proposed framework includes a bespoke problem formulation for manoeuvre prediction, a novel transformer-based prediction model, and a tailored training method for multimodal manoeuvre and trajectory prediction. The performance of the framework is evaluated using three public highway driving datasets, namely NGSIM, highD, and exiD. The results show that our framework outperforms the state-of-the-art multimodal methods in terms of prediction error and is capable of predicting plausible manoeuvre and trajectory modes.

\end{abstract}


\section{Introduction}

Predicting the behaviour of other road users, including vehicles, is crucial for safe and informed decision-making in automated/autonomous vehicles (AVs). A vehicle's future behaviour is inherently multimodal and nondeterministic due to multiple degrees of freedom in a typical driving scenario and the inherent lack of knowledge about the intentions and reactions of other road users. An AV shall consider all plausible future behaviour modes (hereinafter referred to as modes)  of the nearby vehicles to effectively assess and manage the risks and enhance driving safety and efficiency. To this end, several multimodal prediction approaches have been proposed in the literature.

A category of multimodal approaches in the literature predicts multiple plausible modes, each represented by a single manoeuvre type and its corresponding trajectory~\cite{Deo_2018, Mersch2021, Chen2022}. Such an approach does not provide a comprehensive picture of the future modes of a vehicle since a single manoeuvre type can have multiple behaviour realisations depending on the driving styles and interactions with other vehicles. For example, aggressive drivers execute a left lane change differently from conservative drivers in terms of the start time and duration of the manoeuvre. In addition, the future behaviour of a vehicle within a prediction horizon can include a sequence of more than one manoeuvre type.

Another category of existing studies predicts multiple modes, each represented by a trajectory, without predicting the manoeuvre of vehicles. There are generative and discriminative  approaches under this category. Generative approaches rely on estimating the probability distribution function (PDF) of the trajectory using generative models such as  Conditional Variational Auto Encoders (CVAEs) 
 in ~\cite{Salzmann2020, Lee2017} or Generative Adversarial Networks (GANs) in ~\cite{Zhao2019, Gupta2018}. They then sample the PDF to generate multiple trajectories. However, these approaches often require many samples to cover all plausible modes, especially those with lower probabilities, which is not ideal from a planning perspective. On the contrary, discriminative approaches assume a predetermined number of modes and predict one trajectory for each mode along with its likelihood~\cite{Liu2021, Huang2022}. The training of these approaches relies on designing a mode selection method to assign each ground truth mode to one of the prediction modes. Poor mode selection methods can lead to convergence problems such as mode collapsing~\cite{Liu2021, Zhang2021} or implausible modes~\cite{Rhinehart2018, Greer2021}. In the former, multiple predicted trajectories converge to a single plausible mode in the driving scene, whereas in the latter, some predicted trajectories do not conform to traffic or scene context, such as off-road or invalid collision predictions.

To address the aforementioned limitations, we propose a novel Multimodal Manoeuvre and Trajectory Prediction (MMnTP) framework. Firstly, we provide a new formulation of manoeuvre prediction based on a vector representation of manoeuvres, which includes a sequence of manoeuvre types and transition times during the prediction window. To increase the plausibility of the predictions, constraints are introduced on the manoeuvre types and the number of allowed manoeuvre changes within the prediction horizon. We then propose a multimodal discriminative manoeuvre prediction model using this new formulation. The proposed framework employs state-of-the-art transformer neural networks~\cite{Vaswani2017} augmented by manoeuvre-specific heads to predict multiple trajectories conditioned on the predicted manoeuvre vectors. To train the model, a novel multimodal manoeuvre prediction loss function and a mode selection method are proposed based on the predicted types and timings of manoeuvres. Our framework is evaluated in highway driving scenarios using three public trajectory datasets, namely NGSIM~\cite{ngsim}, highD~\cite{highD}, and exiD~\cite{exiDdataset}. Our contributions can be summarised as follows:
\begin{itemize}
    \item A new formulation of manoeuvre prediction, which allows estimating a sequence of manoeuvre types and transition times between them.
    \item A novel transformer-based model to predict multimodal manoeuvres and their corresponding trajectories. 
    \item A tailored multimodal training method using a new multimodal manoeuvre loss function and mode selection method.
    \item Comprehensive and comparative performance evaluation of MMnTP using well-known benchmark datasets and state-of-the-art multimodal models.
\end{itemize}

\section{Related Works}\label{sec:relworks}
Anticipating the future behaviour of other vehicles has been one of the main focuses of research in intelligent vehicles for the last few decades. Early works, reviewed in~\cite{survey_motion_risk}, use physics-based kinematic models to predict the short-term future motion. In recent years, learning-based methods, reviewed in~\cite{mozaffari2022Survey, Liu2021s}, have become popular due to their increased prediction accuracy and extended prediction horizon (i.e., more than 3 seconds). In this section, we specifically review two subsets of recent learning-based methods that correlate with our proposed approach.
\subsection{Manoeuvre-based Multimodal prediction}
These methods rely on early recognition of the manoeuvre intention of vehicles as prior knowledge for trajectory prediction. The authors in~\cite{Deo_2018, Deo20118iv} proposed encoder-decoder neural networks for manoeuvre-based multimodal trajectory prediction. In both papers, the prediction modes include six lateral/longitudinal manoeuvres for highway driving scenarios. The decoder network estimates the probability of each mode and predicts one trajectory per mode. In~\cite{Chen2022}, a three-layer hierarchical network consisting of Long Short-Term Memory (LSTM) and attention mechanism is used to encode the vehicles' motion, and their social interaction with other vehicles. Then, a decoding mechanism similar to that in~\cite{Deo20118iv} is used to predict manoeuvre-based trajectories. In~\cite{Messaoud2021} an attention-pooling architecture with specialised heads per manoeuvre type is introduced. In~\cite{galceran2017multipolicy} the distributions over multiple future policies (i.e., manoeuvre types) of nearby vehicles are estimated, and then the policies are sampled to plan for the Ego vehicle accordingly. Similar to our paper, Mersch~\textit{et al.}~\cite{Mersch2021} predicted the sequence of manoeuvre types of vehicles at one-second time steps as an auxiliary feature to predict a single trajectory output. Their model can generate numerous manoeuvre sequences, but most of them are unrealistic in real-world driving as drivers don't frequently change manoeuvres. We address this problem by limiting the number of manoeuvre changes in our predictions. Additionally, we introduce a multimodal framework for manoeuvre and trajectory prediction, in contrast to Mersh et al.'s single-modal model.

\subsection{Dynamic Multimodal Prediction}
In these methods, multiple trajectories are predicted without using pre-defined manoeuvre types. Such studies broadly fall into the two following sub-categories:

\subsubsection{Generative Models} They estimate the multimodal probability distribution of future trajectories conditioned on past observation. These models rely on sampling to generate multiple future trajectories. In~\cite{Zyner2020, Mercat2020} mixture density model is applied on top of a Recurrent Neural Network (RNN) to model the multimodal distribution. Conditional Variational Auto Encoders (CVAEs) are commonly used to explicitly encode the aforementioned distribution~\cite{Lee2017, Zhao2019, Salzmann2020, Cui2021, Yuan2021}. However, several studies use Generative Adversarial Networks (GAN) to implicitly model the multimodality~\cite{Gupta_2018_CVPR, Sadeghian_2019_CVPR, Vinet2019}. In general, generative methods require a very large number of samples to cover all prediction modes, especially the ones with low probability that are not necessarily unimportant. In addition, such models do not provide a likelihood value for each predicted mode, which is necessary for risk assessment in AVs. 

\subsubsection{Discriminative Models} They directly regress multiple future trajectories and their probabilities. In~\cite{Cui2019} a convolutional neural network is applied on a rasterised bird-eye view representation of the driving environment to predict multiple futures. During training, a mode selection method is defined to select the matching mode to the ground truth for each data sample. The winning mode receives the weight updates through gradient descent optimisation. The authors argue that modelling the multimodal behaviour highly depends on the ability of the mode selection method to distinguish between different modes in a driving scenario. Several studies~\cite{Zhang2021, Kim2021, Messaoud2021, Huang2022, Liu2021, Zhang2021} adopt similar training strategies while attempting to increase compliance with the driving environment and diversity of the predictions. In~\cite{Zhang2021} a scoring and selection layer is introduced to rank the modes and reject the near-duplicates. Kim~\textit{et al.}~\cite{Kim2021} propose a lane-aware feature extraction to predict trajectories in compliance with lanes. To increase the diversity of models, Haung et al.~\cite{Huang2022} use transformers with multimodal attention heads where each mode is assigned to a group of attention heads. Liu et al~\cite{Liu2021} propose a region-based training algorithm for stacked transformers to improve diversity. They group prediction modes based on plausible driving regions. In~\cite{Zhang2021} a set-based neural network architecture is used to predict trajectories from a set of map adaptive proposals. In this paper, we use a dynamic discriminative multimodal prediction approach. However, unlike existing studies, multiple future modes are predicted in manoeuvre space, enabled by vector representation of manoeuvres. Also, a mode selection method is introduced to differentiate between future modes based on the predicted manoeuvre vectors.

\section{Proposed Framework}\label{sec:method}
This section describes the proposed \textbf{M}ultimodal \textbf{M}anoeuvre \textbf{a}nd \textbf{T}rajectory \textbf{P}rediction (\textbf{MMnTP}) framework. First, we define the multimodal manoeuvre and trajectory prediction problem and notations in Section~\ref{sec:prob}.
We then describe the transformer-based prediction model and its main components in Section~\ref{sec:model}. Finally, the training process is explained in Section~\ref{sec:train}.

\subsection{Problem Definition}\label{sec:prob}
We define the problem of multimodal manoeuvre and trajectory prediction as the estimation of multiple future modes, each represented by a sequence of manoeuvre types which we call a manoeuvre vector $\hat{M}_i$, a corresponding trajectory $\hat{Y}_i$ and a probability of occurrence $p_i$, where $i\in\{1,.., N\}$ and $N$ is the number of prediction modes. The prediction is carried out for $T_{pred}$ future time steps based on the observation $O$ of a target vehicle (TV) and its surrounding vehicles (SVs) during the past $T_{obs}$ time steps. The predicted manoeuvre vector of mode $i$ is defined as $\hat{M}_i=\{\hat{m}_{i,t}\}_{t=1}^{T_{pred}}$, and its corresponding trajectory is defined as $\hat{Y}_i=\{\hat{y}_{i,t}\}_{t=1}^{T_{pred}}$, where $\hat{y}_{i,t} =(\hat{y}^{long}_{i,t},\hat{y}^{lat}_{i,t})$ is the predicted x-y location (in meters) of mode $i$ at time-step $t$. The predicted manoeuvre type, $\hat{m}_{i,t}$, is the intention of the TV when moving from location $\hat{y}_{i,t-1}$ to $\hat{y}_{i,t}$. Similarly, $M=\{m_t\}_{t=1}^{T_{pred}}$ and $Y=\{y_t\}_{t=1}^{T_{pred}}$ denote the ground-truth future manoeuvre vector and trajectory, respectively. All manoeuvre types belong to a countable pre-defined manoeuvre set $\bar{M}$. In highway driving, three manoeuvre types are defined as Lane Keeping (LK), Right Lane Change (RLC), and Left Lane Changes (LLC).

\begin{figure}[!t] 
\centering
\includegraphics[width=\linewidth]{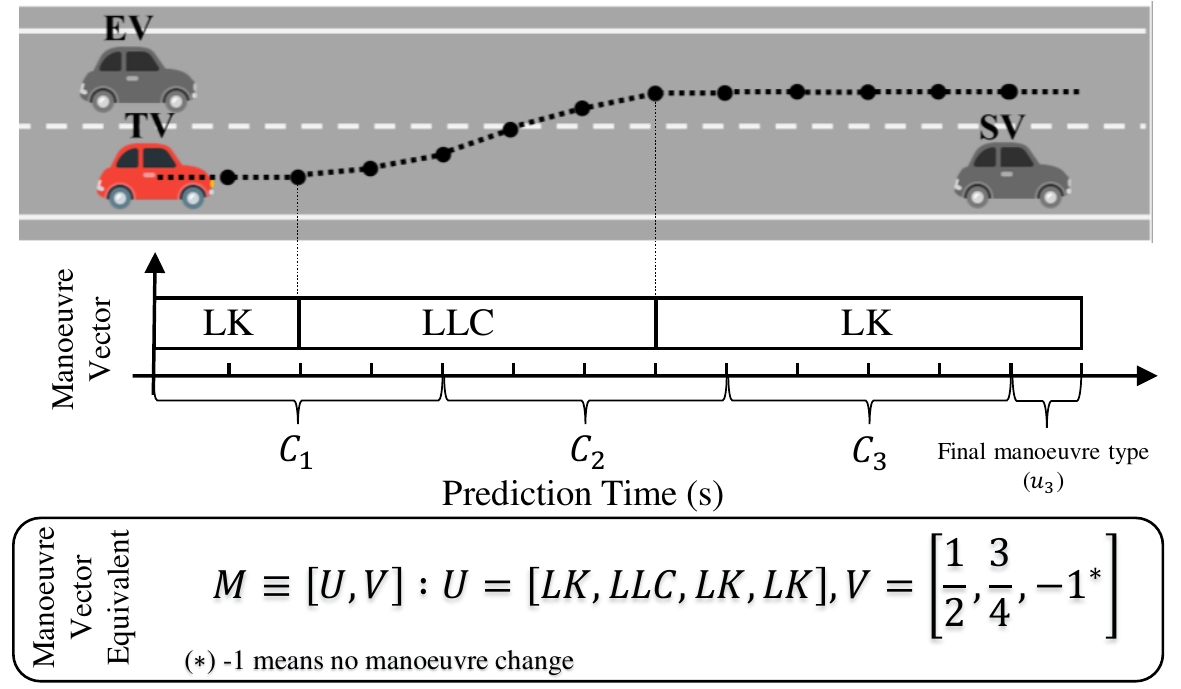}
\caption{An example of a manoeuvre vector with three change periods: $\{C_1, C_2, C_3\}$. The following terminology is used throughout the paper. EV: Ego Vehicle, TV: Target Vehicle, SV: Surrounding Vehicle.}
\label{man_fig}
\end{figure}

We assume that vehicles do not change their manoeuvre in less than $T_{change}$ time steps. Therefore the prediction horizon, $T_{pred}$, can be divided into $C=\lceil \frac{T_{pred}}{T_{change}}\rceil$ change periods. A manoeuvre vector is written as $M \equiv \{U, V\}$, where  $U = \{u_i\}_{i=0}^{C}$ and $V = \{v_i\}_{i=1}^{C}$ are the manoeuvre type and manoeuvre transition time vectors, respectively. Precisely, $u_i \in \bar M$ is the manoeuvre type at the beginning of the $(i+1)$-th change period and $v_i \in [0,1]$ is the transition time from  type $u_{i-1}$ to $u_i$. The transition times are determined by measuring the duration from the beginning of each manoeuvre change period and then normalizing them to the duration of the change period. If $u_{i-1}=u_i$, the transition time, $v_i$, will be ignored (its value is set to $-1$).  Fig.~\ref{man_fig} demonstrates an example of a vehicle trajectory and its respective manoeuvre vector.


\subsection{Prediction Model}\label{sec:model}
\begin{figure*}[!t]
\centering
\includegraphics[width=\textwidth]{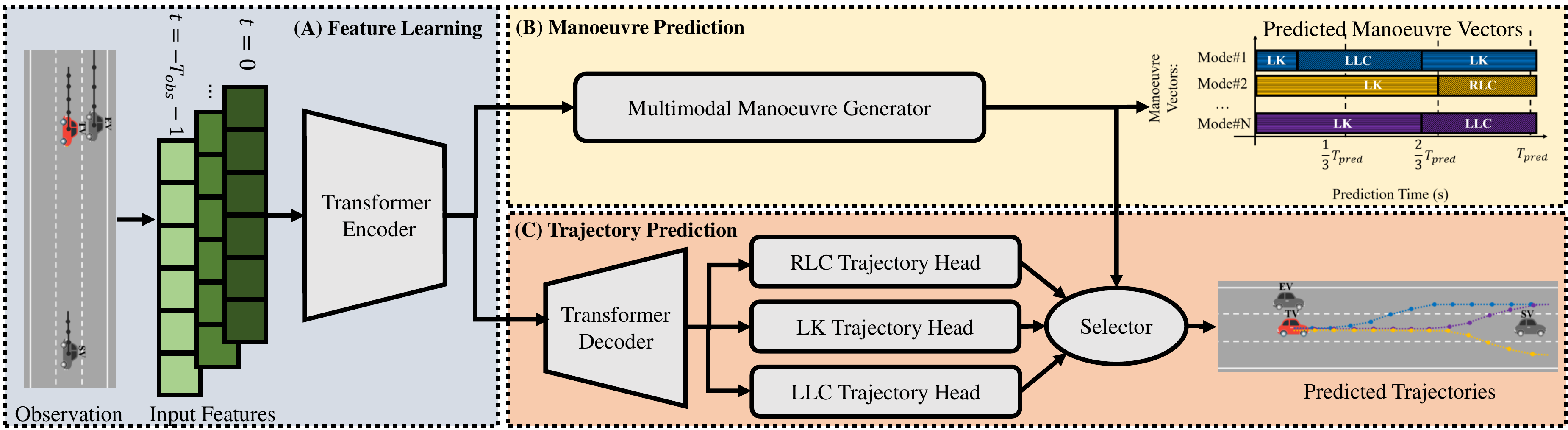}
\caption{An overview of the MMnTP prediction model.}
\label{model_fig}
\end{figure*}

The prediction model includes three main components. First, a transformer encoder extracts latent features from a sequence of interaction-aware inputs (Section~\ref{sec:feature}). Then, a multimodal manoeuvre generator estimates multiple manoeuvre vectors and their likelihoods (Section~\ref{sec:man}). Finally, a trajectory predictor uses the learnt features and estimated manoeuvre vectors to predict their associated trajectory (Section~\ref{sec:traj}). Fig.~\ref{model_fig} illustrates the architecture of the MMnTP.

\subsubsection{Feature Learning}\label{sec:feature}
The input data to the prediction model is the track history of the TV and its SVs, and the position of lane markings during an observation window of $T_{obs}$ time steps. We consider eight SVs including the preceding and following vehicles and the three closest vehicles in each of the adjacent lanes. In~\cite{Wirthmuller2021}, a systematic study has been performed to identify the most relevant features for behaviour prediction in highway driving scenarios, which are also validated by our empirical studies in~\cite{mozaffari2022}. The same set of features is used in this study. These features describe the motion of the TV (e.g., its lateral acceleration), its interaction with SVs (e.g., the relative longitudinal velocity of the TV with respect to the following vehicle), and the driving environment (e.g., the existence of right/left lanes).

A transformer encoder model introduced in~\cite{Vaswani2017}, is used to encode the input feature sequence. Within the encoder model, the input features are embedded into vectors of $512$ dimensions. Then, a sinusoidal positional encoding (proposed in~\cite{Vaswani2017}) is added to the embedded input to specify the order of each input in the sequence.  Finally, eight heads of self-attention are applied to the embedded data, followed by a normalisation layer and a linear layer of size $128$.

\subsubsection{Manoeuvre Prediction} \label{sec:man}
The multimodal manoeuvre generator takes as input the hidden representation learnt by the encoder and predicts $N$ manoeuvre vectors and their probabilities. The model is a two-layer fully-connected neural network with $256$ hidden neurons and ReLU activation function. The output layer contains $N$ neurons for the estimation of mode probabilities, $N\times (C+1)\times3$ neurons for the classification of manoeuvre types, and $N\times C$ neurons for the regression of manoeuvre change times. Note that $N$ is the number of prediction modes and $C$ is the number of change periods, as explained in Section~\ref{sec:prob}. 

\subsubsection{Trajectory Prediction} \label{sec:traj}
The trajectory prediction component includes a transformer decoder network, introduced in~\cite{Vaswani2017}, and manoeuvre-specific heads. Similar to the transformer encoder, the decoder network includes embedding, positional encoding, multi-head self-attention, normalisation layers, and linear layers. In addition, the decoder network benefits from a multi-head cross-attention mechanism to find relevancy between the decoder output and the encoded latent representation. We refer interested readers to~\cite{Vaswani2017} for more details about the transformers. Similar to the encoder network, we use an embedded size of $512$, eight attention heads, and a linear layer of size $128$ in the transformer decoder.

The output of the transformer decoder is fed to manoeuvre-specific heads to predict each coordinate of a trajectory based on the estimated manoeuvre type. For the application of the framework in highway driving scenarios, three heads are considered corresponding to RLC, LK, and LLC manoeuvre types. 
Each head is a linear layer and outputs the parameters of a bi-variate Gaussian distribution (i.e., the x-y means, variances, and correlation coefficient) for each prediction time step. During inference, the future trajectory is predicted using the heads that correspond to the sequence of manoeuvre types within the predicted manoeuvre vector. The mean values of the estimated Gaussian distributions are reported as the predicted coordinates of a trajectory.
 

\subsection{Training Process} \label{sec:train}
We use end-to-end learning to predict multiple manoeuvre vectors, their corresponding trajectories and likelihoods from a sequence of input features. The training loss $L$ consists of a trajectory prediction part $L^{traj}$ and a multimodal manoeuvre prediction part $L^{man}$.
\begin{equation}
    L = L^{traj} + L^{man}.
\end{equation}

The term $L^{traj}$ aims to minimise the negative log-likelihood (NLL) of the estimated bi-variate Gaussian distributions evaluated at each time step of the future ground truth trajectory $Y=\{y_t\}_{t=1}^{T_{pred}}$. We use the teacher-forcing technique in optimising the model's parameters~\cite{Williams1989}. Therefore, for prediction at time step $t$, the ground truth manoeuvre vector $M_{1:t}=\{m_j\}_{j=1}^t$ and the ground truth trajectory $Y_{1:t-1}=\{y_j\}_{j=1}^{t-1}$ at previous time steps are used. Thus, the trajectory prediction loss can be written as:
\begin{equation}
\begin{aligned}
    &-\log\Bigl(f_{\hat Y}(y_{1}|m_{1}, O, \hat{\Omega}_1)\prod_{t=2}^{T_{pred}} f_{\hat Y}(y_{t}|M_{1:{t}},Y_{1:{t-1}}, O, \hat{\Omega}_t)\Bigl),
\end{aligned}
\end{equation}
where $f_{\hat Y}$ is a bivariate Gaussian distribution parameterised by $\hat{\Omega}_t$ at time-step t.

Inspired by the Multiple Trajectory Prediction loss (MTP)~\cite{Cui2019}, we propose a Multiple Manoeuvre Prediction (MMP) loss in the manoeuvre space. Specifically, a novel mode selection method is defined as follows:
\begin{equation}
    n^* =  \underset{n\in \{1,...,N\}}{\argmin} L^{\hat{U}}_n,
\end{equation}
where $L^{\hat{U}}_n$ is the NLL loss for the classification of manoeuvre types in mode $n$ that can be read as:
\begin{equation}
    L^{\hat{U}}_{n} = -\log \prod\limits_{c=0}^{C} q_{n,c}, 
\end{equation}
where $q_{n,c}$ is the estimated probability of ground truth manoeuvre type $u_c$ for mode $n$. Once the winning mode $n^*$ is identified, the manoeuvre loss can be calculated as:
\begin{equation}
    L^{man} = L^{p} +  L^{\hat{U}}_{n^*} +  L^{\hat{V}}_{n^*},
\end{equation}
where $L^{p} = -\log p_{n^*}$ is the NLL of the estimated probability of the winning mode and $L^{\hat{V}}_{n^*}$ is the Euclidean norm between the estimated manoeuvre change timings and their ground truth values for the winning mode that can be read as:
\begin{equation}\label{eq:man_losses}
\begin{aligned}
    & L^{\hat{V}}_{n^*} = \lVert V-\hat V_{n^*}\lVert_2. 
\end{aligned}
\end{equation}

\section{Performance Evaluation}~\label{sec:eval}
In this section, we evaluate the performance of the MMnTP framework in highway driving scenarios. We first explain the selected datasets, the data preprocessing steps, and the evaluation metrics, followed by the implementation details. Then, we compare the performance of the framework with state-of-the-art models. We also provide two ablation studies to analyse the impact of the number of prediction modes and the proposed training process in our framework. Finally, we provide the results of a contingency motion planner using MMnTP prediction data.

\subsection{Dataset and Data Preprocess}
We consider three real-world trajectory datasets for the highway driving scenario evaluation, namely NGSIM~\cite{ngsim}, highD~\cite{highD}, and exiD~\cite{exiDdataset}. The NGSIM dataset includes two different highways in the US with $5$- to $6$- lanes and mild, moderate, and congested traffic conditions. The highD dataset contains different traffic states (e.g., traffic jams) in $2$- to $3$-lane highways in six different locations. NGSIM and highD are the benchmark highway trajectory datasets used in many studies~\cite{Tang2019, Gupta_2018_CVPR, Chen2022}. We create training, validation, and test sets for each dataset with ratios of $70\%$, $10\%$, and $20\%$, following the experiment protocol of~\cite{Tang2019}. The exiD is a new highly interactive dataset containing challenging merging and exiting highway scenarios and is not exploited in multimodal prediction studies yet. We follow the experiment protocol from~\cite{mozaffari2023trajectory} to prepare the train, validation and test set and preprocess the exiD dataset. To this end, we use single-lane merging scenarios from four locations of the exiD dataset and convert their trajectories and lane markings from Cartesian to Frenet coordinates.

Each trajectory in the datasets is automatically labelled with a manoeuvre vector of the same length. If a trajectory crosses a right/left lane marking it is labelled as RLC/LLC for all timesteps before and after crossing until its lateral speed comes to zero. All other timesteps are labelled as lane-keeping.

\subsection{Evaluation Metrics}
\textbf{minRMSE-K} is commonly reported in multimodal trajectory prediction studies~\cite{Gupta2018, Tang2019, Mercat2020}. It is defined as the minimum Root Mean Squared Error (RMSE) among the $K$ selected trajectories for evaluation. In a  model with $N$ prediction modes, $K\leq N$ are the trajectories corresponding to modes with the highest probabilities.

\textbf{meanNLL} is the weighted average NLL (the lower the better) among all prediction modes. The log-likelihood is associated with the probability of observing ground truth trajectory samples given the estimated distribution function. The meanNLL is used to compare multimodal with single modal models (see section~\ref{sec:ablation1}). 

\textbf{CollisionRate, OffroadRate} are the ratios of collisions with SVs and off-road predictions respectively averaged over all predicted modes (the lower the better). A predicted trajectory is considered as a collision or an off-road if at any time-step it overlaps with the bounding box of other vehicles or its centre crosses road borders, respectively. These metrics indicate the feasibility of prediction in terms of future interaction with other vehicles and compliance to map data~\cite{Greer2021}, respectively. 

$\mathbf{\text{\textbf{div}}_K}$ is introduced in this paper as an indicator for the diversity of prediction modes. It is the ratio of pairwise non-overlapping prediction for $K$ prediction modes with the highest probability:
\begin{equation}
        {\text{div}}_K = 1-\dfrac{1}{K(K-1)} \sum\limits_{i=1}^{K} \sum\limits_{j=1,j\neq i}^{K} 
     {\text{overlap}}(i,j)),
\end{equation}
where
\begin{equation}
\text{overlap}(i,j)=
\begin{cases}
    1, & \makecell{\lvert\hat{y}^{lat}_{i, T_{pred}}-\hat{y}^{lat}_{j, T_{pred}}\rvert<2~{\text{m}} \,\,\text{and} \\
    \lvert\hat{y}^{long}_{i, T_{pred}}-\hat{y}^{long}_{j, T_{pred}}\rvert<5~{\text{m}}}\\
    0,              & \text{otherwise},
\end{cases}.
\end{equation} 

\textbf{maxACC-K} is used to evaluate the performance of the multimodal manoeuvre prediction model and is defined as the maximum accuracy among the $K$ manoeuvre prediction vectors with the highest probabilities (the higher the better).

\subsection{Implementation Details}
The prediction model is trained independently on the balanced training set of each dataset. Similar to~\cite{Tang2019}, we use three seconds of observation to predict the next five seconds. We use five frames per second (FPS) for both observation and prediction windows. We consider 2.5 seconds for each manoeuvre intention change period ($C=2$). We use a single layer of transformer encoder-decoder for our evaluations with NGSIM and highD datasets, and two layers for evaluations with the exiD. This is because the exiD dataset has more challenging vehicle interactions. Interested readers are referred to the paper's GitHub page available at \href{https://github.com/SajjadMzf/TrajPred}{https://github.com/SajjadMzf/TrajPred} for further information regarding re-implementing the framework.

\begin{table}
\centering
\caption{Comparision with baselines in terms of  MinRMSE-K at different values of $K$ and prediction horizons}
\label{tab:comparative}
\begin{tblr}{
  row{1-25} = {c},
  cell{2}{1} = {r=13}{},
  cell{16}{1} = {r=6}{},
  cell{22}{1} = {r=3}{},
  cell{2}{2} = {r=6}{},
  cell{8}{2} = {r=4}{},
  cell{12}{2} = {r=2}{},
  cell{14}{2} = {r=2}{},
  cell{16}{2} = {r=4}{},
  cell{20}{2} = {r=2}{},
  cell{22}{2} = {r=2}{},
  hline{1,25} = {-}{0.08em},
  hline{16,22} = {-}{0.03em},
  hline{2} = {-}{},
  hline{8,12,14,20,24} = {2-8}{0.03em},
}
 & K & Model     & 1 s           & 2 s           & 3 s           & 4 s           & 5 s           \\
\begin{sideways}NGSIM\end{sideways}   & 1 & CV          & 0.73          & 1.78          & 3.13          & 4.78          & 6.68          \\
        &   & CS-LSTM~\cite{Deo_2018}  & 0.58          & 1.26          & 2.07          & 3.09          & 3.98          \\
        &   & SAMMP~\cite{Mercat2020}    & 0.51          & 1.13          & 1.88          & 2.81          & 3.67          \\
        &   & STDAN~\cite{Chen2022}     & 0.42          & 1.01          & 1.69          & 2.56          & 4.05          \\
        &  & MFP~\cite{Tang2019}      & 0.54          & 1.16          & 1.9           & 2.78          & 3.83          \\
        &   & MMnTP     & \textbf{0.36} & \textbf{0.96} & \textbf{1.69} & \textbf{2.56} & \textbf{3.55} \\
        & 3 & MATF~\cite{Zhao2019} & 0.66          & 1.34          & 2.08          & 2.97          & 4.13          \\
        &   & S-GAN~\cite{Gupta2018}    & 0.57          & 1.32          & 2.22          & 3.26          & 4.4           \\
        &  & MFP~\cite{Tang2019}      & 0.54          & 1.17          & 1.91           & 2.78          & 3.83          \\
        
        &   & MMnTP     & \textbf{0.26} & \textbf{0.67} & \textbf{1.15} & \textbf{1.73} & \textbf{2.4}  \\
        & 5 & MFP~\cite{Tang2019}      & 0.55          & 1.18          & 1.92           & 2.78          & 3.80          \\
        &   & MMnTP     & \textbf{0.23} & \textbf{0.58} & \textbf{1}    & \textbf{1.49} & \textbf{2.06} \\
        & 6 & SAMMP~\cite{Mercat2020}     & 0.31          & 0.71          & 1.20          & 1.80          & 2.55          \\
        &   & MMnTP     & \textbf{0.22} & \textbf{0.56} & \textbf{0.95} & \textbf{1.42} & \textbf{1.96} \\
\begin{sideways}highD\end{sideways}& 1 & CV          & \textbf{0.09}          & 0.32          & 0.67          & 1.14          & 1.73          \\
        &   & CS-LSTM~\cite{Deo_2018} & 0.19          & 0.57          & 1.16          & 1.96          & 2.96          \\
        &   & STDAN~\cite{Chen2022}   & 0.19          & \textbf{0.27}          & \textbf{0.48}          & \textbf{0.91}          & 1.66          \\
        &   & MMnTP                   & 0.19 & 0.38 & 0.62 & 0.95 & \textbf{1.39} \\
        & 3 & S-GAN~\cite{Gupta2018}  & 0.3           & 0.78          & 1.46          & 2.34          & 3.41          \\
        &   & MMnTP                   & \textbf{0.1}  & \textbf{0.2}  & \textbf{0.34} & \textbf{0.57} & \textbf{0.87} \\
\begin{sideways}exiD\end{sideways}& 1 & CV          & \textbf{0.25}          & 0.63          & 1.19          & 1.92          & 2.82          \\
        &   & MMnTP          & 0.26          & \textbf{0.57}          & \textbf{0.98}          & \textbf{1.50}          & \textbf{2.11}          \\
        & 3 & MMnTP          & \textbf{0.17}          & \textbf{0.39}          & \textbf{0.69}          & \textbf{1.07}          & \textbf{1.52}          \\
        
\end{tblr}
\vspace{-20 pt}
\end{table}
\subsection{Comparison with Baseline Models}
Table~\ref{tab:comparative} reports the performance of the MMnTP framework with six modes ($N=6$), Constant Velocity (CV) predictor, and state-of-the-art multimodal trajectory prediction approaches evaluated on NGSIM, highD, and exiD datasets. The multimodal prediction models are those from the literature that report minRMSE-K on any highway trajectory datasets at least for one value of $K$ in five seconds prediction horizon. These models include manoeuvre-based multimodal approaches like CS-LSTM~\cite{Deo_2018} and STDAN~\cite{Chen2022}, dynamic discriminative multimodal approaches like SAMMP~\cite{Mercat2020}, and generative multimodal approaches like S-GAN~\cite{Gupta2018}, MFP~\cite{Tang2019}, and MATF~\cite{Zhao2019}. To the best of our knowledge, there is no existing multimodal trajectory prediction study on the exiD dataset. We report minRMSE-K of our framework for different values of $K$ from $1$ to $6$.

The results show that our framework outperforms baseline and CV models for all values of $K$ in the five-second prediction horizon. The prediction performance of our framework strictly increases by considering more modes in the evaluation (i.e., increasing $K$). This means that the new modes considered for evaluation add a diversity that lowers the prediction error (i.e., effective diversity). Such diversity is not observed in generative models such as MFP, which has roughly the same minRMSE-K errors for all values of $K$. The CV prediction has a relatively low short-term prediction error, while it is substantially outperformed in the five-second prediction horizon by most learning-based methods. All prediction models, including ours, have lower performance on NGSIM as compared to highD and exiD, which can be linked to high annotation errors of the NGSIM dataset~\cite{ngsim_critic}.

\begin{table}[]
\centering
\caption{Percentage of maximum accuracy of manoeuvre intention prediction among K high probable modes (MaxACC-K) evaluated on balanced NGSIM and highD}
\label{tab:man_perf}
\begin{tabular}{@{}lllllll@{}}
\toprule
Dataset & k=1   & k=2   & k=3   & k=4   & k=5   & k=6   \\ \midrule
NGSIM   & 64.6  & 81.2  & 86.01 & 88.04 & 89.86 & 90.48 \\
highD   & 82.04 & 93.26 & 94.89 & 95.81 & 95.97 & 96.03 \\
exiD   & 79.91 & 89.54 & 93.51 & 95.08 & 95.70 & 95.98 \\
\bottomrule
\end{tabular}
\end{table}

\begin{table}[]
\centering
\caption{Ablation study on impact of number of modes(N) in the MMnTP model on prediction performance}
\label{tab:abb}
\begin{tabular}{@{}lllllll@{}}
\toprule
Metric                     & N & \multicolumn{1}{c}{1 s} & \multicolumn{1}{c}{2 s} & \multicolumn{1}{c}{3 s} & \multicolumn{1}{c}{4 s} & \multicolumn{1}{c}{5 s} \\ \midrule
\multirow{3}{*}{MinRMSE-1} & 1 & \textbf{0.19}           & \textbf{0.36}           & \textbf{0.56}           & \textbf{0.82}           & \textbf{1.19}           \\
                           & 3 & 0.19                    & 0.38                    & 0.62                    & 0.95                    & 1.37                    \\
                           & 6 & 0.21                    & 0.41                    & 0.67                    & 1.01                    & 1.46                    \\ \midrule
\multirow{2}{*}{MinRMSE-3} & 3 & \textbf{0.1}            & \textbf{0.21}           & \textbf{0.36}           & \textbf{0.58}           & \textbf{0.88}           \\
                           & 6 & 0.11                    & 0.21                    & 0.37                    & 0.6                     & 0.91                    \\ \midrule
MinRMSE-6                  & 6 & \textbf{0.07}           & \textbf{0.15}           & \textbf{0.28}           & \textbf{0.46}           & \textbf{0.72}           \\ \midrule
\multirow{3}{*}{meanNLL}      & 1 & \textbf{-5.08}          & \textbf{-4.77}          & -3.79                   & -2.06                   & 0.05                    \\
                           & 3 & -4.99                   & -4.55                   & -3.65                   & -2.38                   & -1.04                   \\
                           & 6 & -5.04                   & -4.63                   & \textbf{-3.79}          & \textbf{-2.67}          & \textbf{-1.49}          \\ \bottomrule
\end{tabular}
\end{table}

\subsection{Manoeuvre Prediction Performance} 
Table~\ref{tab:man_perf} demonstrates the manoeuvre prediction performance of the MMnTP reported on NGSIM, highD, and exiD test sets. The table reports MaxACC-K for $K = {1,2,...,6}$ modes. Particularly for this experiment, we balance the test sets in terms of the type of manoeuvres, since accuracy metrics can provide misleading results on unbalanced datasets. The results show that for $K=6$ the model achieves an accuracy of more than $90\%$, while with $K=1$ it achieves around $64\%$ on NGSIM, $82\%$ on highD and $79\%$ on exiD. We speculate that the lower performance on NGSIM could be related to inaccuracies in manoeuvre labels on this dataset caused by tracking errors which also cascades to lower trajectory prediction performance on this dataset.

\begin{figure}[t!]
\centering
\includegraphics[width=.8\linewidth]{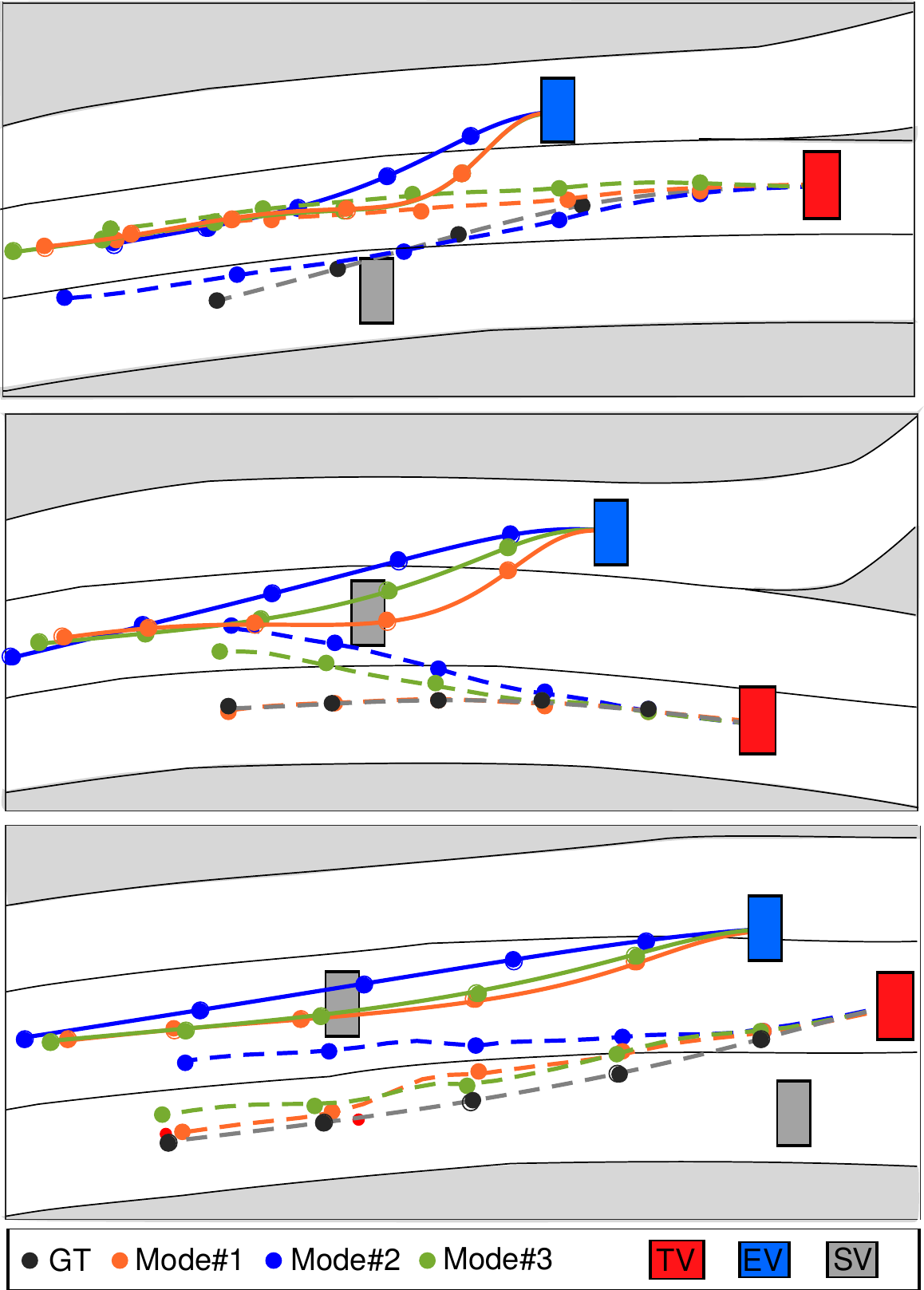}
\caption{Contingency motion planning for MMnTP's multimodal prediction in three different merging scenarios extracted from the exiD dataset. Trajectory predictions and plans are depicted with dashed and continues lines of matching colours, respectively.}
\label{fig: contingencyPlanning}
\centering
\end{figure}

\subsection{Ab. Study 1: Impact of Number of Prediction Modes}\label{sec:ablation1}
We investigate the impact of $N$, the number of prediction modes on the performance of trajectory prediction of our framework. We consider $N={1,3,6}$, where $N=1$ is the single modal variant (e.g., transformer encoder decoder only). Table~\ref{tab:abb} shows the results using minRMSE-K and meanNLL metrics on the validation set of the highD dataset. The results indicate that for minRMSE-1 the single modal variant achieves the lowest error. We speculate this is because single modal approaches are prone to converging to averages of modes which can yield lower RMSE scores. Nonetheless, among all minRMSE-K metrics, the model with 6 modes achieves the best performance. In addition, the multimodal variant with 6 modes achieves the lowest meanNLL in long-term predictions (i.e., 3 to 5-second prediction horizon).

\begin{table} 
\centering
\caption{Ablation study on the impact of proposed training process on accuracy, plausibility, and diversity of predictions}
\resizebox{\columnwidth}{!}{
\begin{tblr}{
  cells = {c},
  hline{1,4} = {-}{0.08em},
  hline{2} = {-}{},
}
\makecell{Training Method} & \begin{sideways}minRMSE-1\end{sideways} & \begin{sideways}minRMSE-2\end{sideways} & \begin{sideways}minRMSE-3\end{sideways} & \begin{sideways}$\text{div}_2$\end{sideways} & \begin{sideways}$\text{div}_3$\end{sideways} & \begin{sideways}\makecell{CollisionRate \\ (\%)}\end{sideways}  & 
\begin{sideways}\makecell{OffroadRate\\(\%)}\end{sideways}  \\
\makecell{MMnTP+MTP}                                                                 & \textbf{1.98}                                    & \textbf{1.62}                                    & 1.62                                    & 0.08                                  & \textbf{0.69}                                  & 9.63                                 & 26.83                                  \\
 \makecell{MMnTP+MMP (ours)}                                                                          & 2.11                                    & 1.71                                    & \textbf{1.52}                                    & \textbf{0.31}                                  & 0.27                                  & \textbf{2.46}                                & \textbf{0.06}                                                                   
\end{tblr}}
\label{tab:abb2}
\end{table}

\subsection{Ab. Study 2: Impact of Training Process}\label{sec:ablation2}
We investigate the impact of the proposed training strategy and the MMP loss function on the performance of our framework. To this end, we train and evaluate a modified version of MMnTP with an alternative well-known training strategy for multimodal trajectory prediction known as MTP loss function~\cite{Cui2019, Liu2021}. The MMnTP+MTP is considered a dynamic multimodal prediction model where the trajectory prediction is not conditioned on estimated manoeuvre. We use the same hyperparameters and trajectory loss function for MMnTP+MTP as in the original framework (i.e., MMnTP+MMP).
The mode selection algorithm for MMnTP+MTP is defined as:
\begin{equation}
    n^* =  \underset{n\in \{1,...,N\}}{\argmin}  \lVert\hat{y}_{n, T_{pred}}-y_{T_{pred}}\rVert_1,
\end{equation}
where $\hat{y}_{n,T_{pred}}$ is the predicted x-y coordinates of mode $n$ at the end of the prediction horizon and $y_{T_{pred}}$ is the corresponding ground-truth value.

We train and evaluate both versions of MMnTP with $N=3$ on the exiD dataset. Table~\ref{tab:abb2} compares the performance of these approaches in terms of prediction accuracy (i.e., minRMSE-K), diversity (i.e., $\text{div}_K$), and plausibility (i.e., CollisionRate and OffroadRate). The results show that the proposed MMnTP+MMP achieves relatively the same accuracy, but much lower collision and offroad predictions, compared to MMnTP+MTP.  Also, note that the high diversity in MMnTP+MTP with three modes (i.e., ${\text{div}}_3$) does not effectively contribute to prediction accuracy because minRMSE-2 and minRMSE-3 are the same for this method.

\subsection{Prediction-based Planning Results}
To assess the impact of the MMnTP on a downstream motion planner, we feed its trajectory predictions to a Model Predictive Control (MPC). We adopt the same planning cost function and parameters as in~\cite{mozaffari2023trajectory}. Similar to~\cite{chen2022scept, cui2021lookout}, we design a contingency planner where $N$ trajectories are planned for the EV based on each of the $N$ predicted modes for a TV. The planning is subject to the constraint that the first control inputs remain the same in all planned trajectories. To increase time efficiency of  prediction-based planning, one surrounding vehicle, which has the most impact on the EV's future trajectory, is selected as the target vehicle in each driving scenario. Fig.~\ref{fig: contingencyPlanning} illustrates the MMnTP-based trajectory predictions and the corresponding EV's planned motion for three merging scenarios extracted from the exiD dataset. The closest left-following vehicle to the EV is selected as the TV in the studied merging scenarios. The results show that MMnTP's diverse and plausible predictions result in various merging plans for the EV.

\section{Conclusion}~\label{sec:conc}
In this paper, we proposed a novel transformer-based multimodal manoeuvre and trajectory prediction (MMnTP) framework using a new formulation of manoeuvre prediction. The performance evaluation on highway driving scenarios indicates that the diverse prediction modes of MMnTP can enhance prediction accuracy as compared to state-of-the-art models. The results show that the multimodal manoeuvre training strategy plays a key role in enhancing the plausibility of predictions. Furthermore, we show that MMnTP can be effectively integrated with a contingency motion planner. In future studies, it is important to extend the current framework to incorporate multi-agent prediction, enabling the prediction of multiple futures for a group of nearby vehicles.



\bibliographystyle{IEEEtran}

\bibliography{references}

\end{document}